\documentclass[10pt,letterpaper]{article}

\usepackage[T1]{fontenc}
\usepackage[utf8]{inputenc}
\usepackage[english]{babel}
\usepackage{amsmath,amssymb,amsthm}
\usepackage{lmodern}
\usepackage{XCharter}
\usepackage[charter]{newtxmath}
\IfFileExists{sourcesans.sty}{\usepackage[scale=.95]{sourcesans}}{\usepackage[scale=.95]{sourcesanspro}}
\usepackage{mathtools,bm}
\usepackage{graphicx,booktabs,array,tabularx,longtable,multirow}
\usepackage{xcolor,tikz}
\usepackage{colortbl}
\usetikzlibrary{arrows.meta}
\usepackage[numbers,square,sort&compress]{natbib}
\usepackage[stretch=20,shrink=25,step=1]{microtype}
\usepackage{enumitem}
\usepackage{caption}
\usepackage{hyperref}
\usepackage{url}
\usepackage{placeins}
\definecolor{ink}{HTML}{202D40}
\definecolor{physics}{HTML}{3D5B87}
\definecolor{neural}{HTML}{647995}
\definecolor{feedback}{HTML}{A44932}
\usepackage{marvosym}

\hypersetup{colorlinks=true,linkcolor=ink,citecolor=physics,urlcolor=physics,
 pdftitle={TelePhysica: Agentic Reconstruction of Physically Interactive Scenes},
 pdfauthor={Ruiqi Shu; Xufeng Zhan; Yuanzhi Liang; Haibin Huang; Chi Zhang; Xuelong Li}}
\setlist{nosep,leftmargin=*}
\newcommand{\R}{\mathbb R}
\newcommand{\dd}{\mathrm d}

\usepackage[textwidth=168mm,top=21mm,bottom=22mm,headheight=14pt,headsep=14pt,footskip=22pt]{geometry}
\usepackage{fancyhdr,titlesec}
\makeatletter
\renewcommand{\normalsize}{%
  \@setfontsize\normalsize{10.2}{13.6}%
  \abovedisplayskip=7pt plus 2pt minus 2pt
  \belowdisplayskip=7pt plus 2pt minus 2pt
  \abovedisplayshortskip=3pt plus 1pt
  \belowdisplayshortskip=4pt plus 1pt}
\renewcommand{\small}{\@setfontsize\small{9.1}{11.7}}
\renewcommand{\footnotesize}{\@setfontsize\footnotesize{8.3}{10.5}}
\makeatother
\normalsize
\renewenvironment{abstract}{%
 \par\begingroup\fontsize{9.4}{12.5}\selectfont\setlength{\parskip}{4pt}%
 \noindent{\sffamily\bfseries\fontsize{10}{12.5}\selectfont\color{physics}Abstract}\par\vspace{3pt}%
}{\par\endgroup\vspace{8pt}}
\titleformat{\section}{\sffamily\bfseries\fontsize{13.8}{17}\selectfont\color{physics}}{\thesection}{.75em}{}
\titleformat{\subsection}{\sffamily\bfseries\fontsize{11.3}{14}\selectfont\color{ink}}{\thesubsection}{.6em}{}
\titleformat{\subsubsection}{\sffamily\bfseries\normalsize\color{ink}}{\thesubsubsection}{.6em}{}
\titlespacing*{\section}{0pt}{15pt plus 3pt minus 2pt}{6pt}
\titlespacing*{\subsection}{0pt}{11pt plus 2pt minus 2pt}{4.5pt}
\titlespacing*{\subsubsection}{0pt}{8pt plus 2pt}{4pt}
\DeclareCaptionFont{reportcaption}{\fontsize{8.9}{11.6}\selectfont}
\fancypagestyle{plain}{\fancyhf{}\fancyfoot[R]{\sffamily\fontsize{8.5}{10}\selectfont\color{physics}\thepage}}
\newcolumntype{Y}{>{\raggedright\arraybackslash}X}
\newcommand{\TeaserCaption}{\textbf{TelePhysica overview.}
(a) Closed-loop recovery from a single photograph $I_0$ refines a procedural scene program via Stage-1 silhouette critique and Stage-2 gravitational equilibrium checks.
(b) Unified stateful mechanics spans $\mathrm{SE}(3)$ rigid contact, $164$-link chains, thin-shell cloth, and 3D tetrahedral solids under surface wrenches $\boldsymbol{\tau}(t) = \mathbf{r}(t) \times \mathbf{F}(t)$ and patch tractions $\mathbf{f}_{\mathrm{ext},a}(t) = w_a \mathbf{F}_{\mathrm{patch}} w(t)$.
(c) Phase-aligned physics and VACE frames across rigid tumbling (A14B RGB), cloth folding (A14B RGB), and solid indentation (A14B Canny).}

\newcommand{\ClosedLoopCaption}{\textbf{Two-stage closed-loop verification versus single-pass open-loop synthesis.}
(a) Single-image reconstruction traces on stacked kitchen sponges and an upholstered chair cushion, showing input photographs, initial and refined renders, normalized foreground mask-difference diagnostics (red: reference-only deficit; cyan: render-only excess), and Stage-2 zero-force nodal drift endpoints. On the sponges, visual refinement reaches $\mathrm{IoU} = 0.55$, whereas raw Stage-2 preflight detects a $68.3\,\mathrm{mm}$ nodal drop over $1.0\,\mathrm{s}$ across an occluded vertical support gap ($0.661\,\mathrm{mm}$ over $4.0\,\mathrm{s}$ after support repair and damped contact relaxation). The chair cushion passes autonomously with $0.979\,\mathrm{mm}$ peak nodal drift over $4.0\,\mathrm{s}$.
(b) Controlled 100-scene tabletop audit comparing first-pass open-loop against closed-loop execution across visual alignment ($2\%$ vs.\ $85\%$), equilibrium verification ($0\%$ vs.\ $81\%$), and full $5.0\,\mathrm{s}$ force rollouts ($0\%$ vs.\ $49\%$), alongside initial Stage-1 ($98/100$) and first Stage-2 preflight ($84/95$) rejection breakdowns.}

\newcommand{\GalleryCaption}{\textbf{Single-image 3D world reconstruction and simulated interactions across constitutive regimes.}
(a--c) Multi-object rigid contact, sliding, and collision cascades pairing input photographs with simulated post-perturbation states.
(d) A $164$-link articulated earphone cable in its initial resting model state and after a prescribed $300\,\mathrm{mm}$ earbud lift.
(e,f) Co-dimensional thin-shell towel corner folding and layered knitted glove cuff peeling.
(g,h) Three-dimensional tetrahedral seat cushion indentation and soft-to-rigid sponge-to-bottle contact transfer.}

\newcommand{\MechanicsCaption}{\textbf{Constitutive proxy ablation, force reversal, and persistent multi-regime state.}
(a) Controlled ablation on a $0.23\,\mathrm{kg}$ microfiber bath towel ($\mu = 0.80$) under $[2.6, 2.6, 0]^\top\,\mathrm{N}$ corner traction ($t = 4.00\,\mathrm{s}$ endpoint metrics; frames shown at $t = 3.96\,\mathrm{s}$): a matched rigid plate slides $\Delta_{xy} = 72.1\,\mathrm{mm}$ in plane with $h_{\mathrm{fold}} = 0\,\mathrm{mm}$, whereas the $121$-node cloth shell pulls $(\Delta x, \Delta y) = (526.7, 564.3)\,\mathrm{mm}$ with $\Delta z_{\max} = 80.4\,\mathrm{mm}$ nodal pleat height, and mirrored $[-2.6, -2.6, 0]^\top\,\mathrm{N}$ opposite-corner traction produces symmetric folding.
(b) Layered-shell XPBD glove response ($F_{\max} = 0.09\,\mathrm{N}$, $t = 0.875\,\mathrm{s}$) comparing zero ($\mathbf{0}$), forward ($+\mathbf{F}$), and reversed ($-\mathbf{F}$) traction.
(c) Prescribed $\Delta z = 300\,\mathrm{mm}$ lift along a $164$-link articulated cable ($t = 0, 2.00, 3.00\,\mathrm{s}$).
(d) Rigid bowling contact rollout alongside a continuous mid-rollout trajectory appending $\mathbf{F} = (45, 0, 0)\,\mathrm{N}$ over $t \in [0.1, 0.2]\,\mathrm{s}$ after an initial $\mathbf{J} = (0, 18, 0)\,\mathrm{N\cdot s}$ impulse without state reset.
(e) Volumetric indentation ($\Delta z_{\min} = -58.2\,\mathrm{mm}$ at $t = 1.04\,\mathrm{s}$) and elastic rebound ($\max |\Delta z| = 0.84\,\mathrm{mm}$ at $t = 4.00\,\mathrm{s}$) on a $768$-tetrahedron seat cushion.
(f) Soft-to-rigid contact where two tetrahedral foam sponges compress against a rigid bottle and slide it $\Delta x = 76.7\,\mathrm{mm}$.}

\newcommand{\VideoComparisonCaption}{\textbf{Phase-aligned comparison of physics-grounded video synthesis against direct image-to-video baselines.}
Rows (a--f) span ceramic cup tumbling, bowl-and-cabbage collision, forward and counterfactual opposite-corner towel pulling, knitted glove cuff peeling, and hybrid sponge-to-bottle pushing. Columns compare the input photograph and aligned physics rollout against unconditioned Direct I2V ($1.3\mathrm{B}$, $\mathrm{A14B}$) and trajectory-conditioned VACE ($1.3\mathrm{B}$ and $\mathrm{A14B}$ under Canny and RGB control) at matched generated frame indices $k \in \{28, 38, 60, 60, 40, 60\}$ (corresponding to simulated physics frames $42, 57, 71, 71, 48, 71$).}

\begin{document}
\thispagestyle{plain}
{\sffamily\fontsize{8.8}{11}\selectfont\color{physics}Technical Report\par}
\vspace{5pt}
{\fontsize{24}{28}\selectfont\bfseries\raggedright\color{ink}TelePhysica: Agentic Reconstruction of Physically Interactive Scenes\par}
\vspace{5pt}
{\sffamily\fontsize{10}{13}\selectfont\color{physics}Turning Any Visual Content into an Interactive Physical World\par}
\vspace{3pt}
{\fontsize{10.5}{14}\selectfont\raggedright Ruiqi Shu* \quad Xufeng Zhan* \quad Yuanzhi Liang \quad Haibin Huang \quad Chi Zhang \quad Xuelong Li\textsuperscript{\Letter} \quad\par}
{\fontsize{9}{12}\selectfont\raggedright Institute of Artificial Intelligence, China Telecom (TeleAI) \par}
\vspace{3pt}
{\color{physics}\hrule height .6pt}
\vspace{7pt}

\begin{abstract}
\noindent
Direct image-to-video diffusion models entangle pixel synthesis with implicit latent motion, precluding persistent 3D mechanical equilibrium, mid-rollout intervention, and counterfactual force control. Conversely, single-pass programmatic 3D reconstruction from an uncalibrated photograph leaves monocular depth and support clearance underconstrained, causing reconstructed scenes to interpenetrate or collapse under gravity. We present \textbf{TelePhysica}, a closed-loop framework that reconstructs an executable, multi-regime physical world from a single photograph to drive stateful 3D interaction and trajectory-grounded video synthesis. A vision-language coding agent synthesizes separable visual meshes and declarative physical models within a two-stage verification loop that pairs pixel-aligned silhouette critique with a zero-force gravitational equilibrium preflight. The verified world compiles into a unified solver coupling rigid bodies, articulated chains, co-dimensional cloth shells, and 3D tetrahedral elastodynamics, whose simulated rollouts condition pretrained video diffusion models through pixel-aligned structural streams. On a 100-scene single-image benchmark, closed-loop execution raises equilibrium-verified world recovery from $0\%$ in single-pass synthesis to $81\%$. Controlled constitutive ablations show over an order-of-magnitude larger localized corner displacement and out-of-plane pleating in finite-element cloth than in matched rigid-plate proxies. Conditioning video diffusion on simulated trajectories increases dynamic motion-mask overlap by $2{\times}\text{--}5{\times}$ and preserves motion direction under counterfactual force reversals where direct image-to-video models fail.
\end{abstract}

\noindent\begin{minipage}{\textwidth}
\centering
\captionsetup{hypcap=false}
\includegraphics[width=0.8\textwidth]{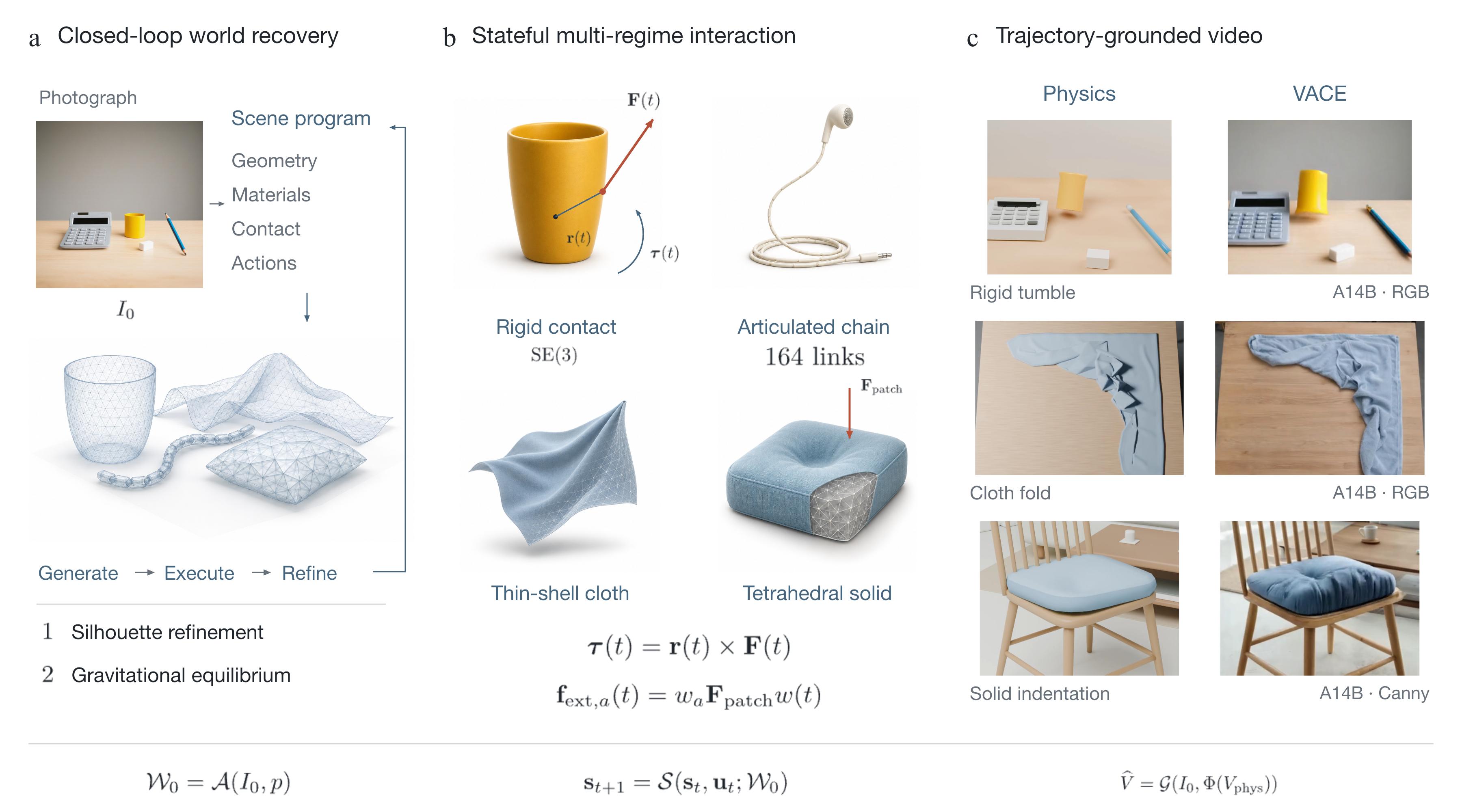}
\captionof{figure}{\TeaserCaption}
\label{fig:teaser}
\end{minipage}
\let\thefootnote\relax\footnotetext{$^*$ denotes equal contributions. This work was completed during their internship at TeleAI. Xuelong Li is the corresponding author. }
\let\thefootnote\relax\footnotetext{ Homepage:  https://interactive4d.vercel.app/ }

\clearpage

\section{Introduction}

Predicting how a scene in a single photograph responds to physical interaction---tipping a ceramic mug near its rim, tugging one corner of a bath towel, or indenting an upholstered cushion---requires reasoning over 3D geometry, frictional contact, and material constitutive laws. Sequential and counterfactual interventions further demand an explicit mechanical state rather than a single synthesized clip: reversing a traction vector, shifting a contact point, or applying a mid-rollout impulse alters lever-arm moments, unilateral support reactions, membrane buckling, and volumetric elastic recovery.

Existing generative and inverse-graphics paradigms face two complementary bottlenecks in interactive physical modeling. First, trajectory- and prompt-conditioned video diffusion models inject text cues, 2D strokes, or sparse point tracks directly into latent denoisers \cite{yin2023dragnuwa,wang2024motionctrl,gillman2025force,wang2025physctrl}. Lacking a persistent 3D mechanical state, these models often freeze deformable surfaces, actuate the wrong object region, or violate momentum balance under counterfactual force reversals. Second, feed-forward 4D synthesis and open-loop inverse-graphics pipelines recover static assets or fixed neural animations \cite{bahmani2024_4dfy,chu2024dreamscene4d,sun2023_3dgpt,hu2024scenecraft}, while monocular projection leaves depth ordering and support clearance ambiguous along viewing rays. Placed directly under gravity without physical verification, single-pass reconstructions fail through non-manifold geometry, initial collider overlap, or support collapse.

TelePhysica resolves both bottlenecks by decoupling 3D world reconstruction, constitutive dynamics, and photorealistic video synthesis while linking them through executable verification gates (Figure~\ref{fig:teaser}). From a single photograph $I_0$ and scene description $p$, a vision-language coding agent synthesizes procedural visual meshes and declarative physical models inside a two-stage loop. Stage~1 aligns rendered silhouettes and interior contours with $I_0$ via pixel-level mask-difference critique; Stage~2 compiles candidate colliders and finite-element meshes into a dynamical solver and enforces a zero-force gravitational equilibrium gate, rejecting floating or interpenetrating layouts that appear valid in 2D projection. Verified worlds then evolve in a unified checkpointable solver spanning rigid bodies, articulated chains, thin-shell cloth, and 3D tetrahedral solids under calibrated surface wrenches, distributed patch tractions, impulses, or prescribed trajectories. Finally, rendered 3D rollouts condition pretrained video generators via structural edge, shaded RGB, or motion-reference streams \cite{wan2025wan21,jiang2025vace,videoxfun2025,minimax2025h3}.

Experiments across single-image reconstruction, continuum mechanics, and generative video synthesis establish three principal results:
\begin{itemize}[topsep=3pt,itemsep=3pt]
\item \textbf{Closed-loop physical world recovery.} On a 100-scene single-image benchmark, two-stage verification lifts equilibrium-verified reconstruction from $0\%$ in single-pass open-loop synthesis ($2\%$ visual pass) to $81\%$ in a median of $7$ scene submissions, while blocking false visual-only acceptance on occluded deformable stacks.
\item \textbf{Multi-regime constitutive fidelity.} Holding geometry, mass, friction, and applied force fixed, co-dimensional finite-element cloth produces $> 10\times$ larger localized corner travel and $8\text{--}11\,\mathrm{cm}$ out-of-plane pleating than a matched rigid-plate proxy, while coupled 3D elastodynamics captures sub-millimeter foam rebound and soft-to-rigid momentum transfer.
\item \textbf{Trajectory-grounded video synthesis.} Conditioning video diffusion on simulated 3D rollouts raises dynamic motion-mask intersection-over-union (IoU) from $0.17\text{--}0.27$ in unconditioned image-to-video baselines to $0.48\text{--}0.87$, reduces centroid trajectory error by $4{\times}\text{--}20{\times}$, and maintains $> 0.90$ directional cosine alignment under counterfactual force reversals where unconditioned models yield negative correlation.
\end{itemize}

\section{Related Work}

\subsection{Programmatic and Agentic 3D Scene Synthesis}

Representing 3D environments as executable graphics programs provides interpretable hierarchy, separable object instances, and parametric editability. Language-guided layout and procedural modeling systems---including LayoutGPT \cite{feng2023layoutgpt}, 3D-GPT \cite{sun2023_3dgpt}, SceneCraft \cite{hu2024scenecraft}, and Holodeck \cite{yang2024holodeck}---translate text specifications into spatial arrangements or Blender scripts, while BlenderAlchemy \cite{huang2024blenderalchemy} and Thinking-in-Blender \cite{he2026thinkinginblender} close a visual render loop to refine materials and geometry. Programmatic world representations also support symbolic physical reasoning in Code-as-Worlds \cite{wang2026codeasworlds} and VisPhyWorld \cite{liang2026visphyworld}, and 4DCodeBench \cite{shen2026_4dcodebench} evaluates coding agents on reconstructing observed dynamic trajectories from monocular videos. Complementarily, distributed model cooperation \cite{an2026aiflow,shao2026aiflownetworkedge} and structure-aware static-scene delivery \cite{yuan2026enhancing} underscore the efficiency of compact, modular scene abstractions. Rather than fitting a single observed trajectory from multi-frame video, TelePhysica reconstructs an equilibrium-verified physical world from a single static photograph to support diverse post-reconstruction force interventions, mid-rollout continuation, and counterfactual queries.

\subsection{Interactive Physical World Modeling Across Material Regimes}

Coupling 3D scene representations with multi-body and continuum solvers \cite{todorov2012mujoco,macklin2016xpbd,coumans2016pybullet} grounds visual synthesis in Newtonian mechanics. Feed-forward 4D generation and score-distillation pipelines \cite{bahmani2024_4dfy,chu2024dreamscene4d} synthesize plausible object motion but bake deformation into neural fields without explicit contact colliders. Differentiable continuum twins---including PAC-NeRF \cite{li2023pacnerf}, PhysGaussian \cite{xie2024physgaussian}, PhysDreamer \cite{zhang2024physdreamer}, Spring-Gaus \cite{zhong2024springgaus}, PIE-NeRF \cite{qiu2024feature}, PhysMotion \cite{tan2024physmotion}, and Phys4DGen \cite{lin2025phys4dgen}---embed material-point, spring-mass, or finite-element dynamics into radiance fields or 3D Gaussian splats, generally assuming multi-view captures or isolated foreground objects. For single-view interaction, PhysGen \cite{liu2024physgen} models planar rigid dynamics in image space, whereas PhysGen3D \cite{chen2025physgen3d}, WonderPlay \cite{li2025wonderplay}, RealWonder \cite{liu2026realwonder}, PerpetualWonder \cite{zhan2026perpetualwonder}, and Fluid-Gen-Zero \cite{huang2026fluidgenzero} lift single photographs into 3D meshes, Gaussian particles, or fluid emitters for action-conditioned rollouts. TelePhysica complements particle-based lifting by compiling agent-synthesized meshes into a unified constraint and finite-element solver that couples six-degree-of-freedom rigid bodies, high-link articulated chains, thin-shell cloth, and 3D tetrahedral solids behind a deterministic checkpoint interface.

\subsection{Trajectory-Conditioned and Aligned Video Generation}

Motion control in video diffusion transformers \cite{yang2024cogvideox,kong2024hunyuanvideo,wan2025wan21} spans latent trajectory injection and dense structural conditioning. VAST 1.0 \cite{zhang2024vast} decouples storyboard layout planning from pixel generation, whereas DragNUWA \cite{yin2023dragnuwa}, MotionCtrl \cite{wang2024motionctrl}, DragAnything \cite{wu2024draganything}, Tora \cite{zhang2024tora}, Force Prompting \cite{gillman2025force}, and PhysCtrl \cite{wang2025physctrl} drive diffusion denoisers with sparse point tracks or local force cues. Without an external 3D contact state, however, out-of-plane rotations, multi-body collisions, and layered cloth disocclusions remain susceptible to silhouette drift and directional ambiguity. Structural video adapters such as VACE \cite{jiang2025vace,videoxfun2025} inject frame-aligned contour or RGB tensors alongside reference frames, while MiniMax-H3 Ref2VA \cite{minimax2025h3} generates joint audio-video streams from paired appearance and motion references. Orthogonal post-training frameworks align generative trajectories with physical plausibility and spatiotemporal preference rewards \cite{liang2026integrating,liang2026teleboost,ni2026seeing,liu2026trust,li2026rethinking,li2026credit,li2026rats}. TelePhysica supplies solver-verified 3D rollouts as conditioning streams for pretrained generators and characterizes how model scale and control density govern trajectory fidelity versus material realism.

\section{Method}

TelePhysica maps an input photograph $I_0 \in \R^{H \times W \times 3}$ and scene description $p$ to an executable physical world $\mathcal{W}_0$ and trajectory-grounded video rollouts $\widehat{V}$ through three stages (Figure~\ref{fig:teaser}): closed-loop inverse graphics with gravitational equilibrium verification (\S\ref{sec:method_agent}), unified multi-regime elastodynamics under calibrated perturbations (\S\ref{sec:method_physics}), and deformable surface skinning coupled with structural video diffusion (\S\ref{sec:method_video}). All stages share a right-handed world frame with $+Z$ opposing gravity $\mathbf{g} = (0, 0, -9.81)^\top\,\mathrm{m/s^2}$, SI units ($\mathrm{m}$, $\mathrm{kg}$, $\mathrm{s}$, $\mathrm{N}$, $\mathrm{Pa}$), and scalar-first unit quaternions $\mathbf{q} = (w, x, y, z) \in S^3$ with rotation matrices $\mathbf{R}(\mathbf{q}) \in \mathrm{SO}(3)$ (Appendix~\ref{app:formulation}).

\subsection{Closed-Loop Inverse Graphics and Physical Equilibrium Verification}
\label{sec:method_agent}

Reconstructing an interactive 3D world from an uncalibrated view $I_0$ requires recovering camera calibration $\mathcal{M}$, separable visual meshes $\mathcal{O} = \{o_i\}_{i=1}^N$, collision primitives and finite-element rest topologies $\mathcal{C}$, mobility roles $\rho_i \in \{\text{dynamic}, \text{kinematic}, \text{static}, \text{visual-only}\}$, and a statically stable initial mechanical state $\mathbf{s}_0$:
\begin{equation}
\mathcal{W}_0 = \bigl(\mathcal{O}, \mathcal{C}, \mathcal{M}, \mathbf{s}_0\bigr) = \mathcal{A}(I_0, p).
\label{eq:world_package}
\end{equation}
Because perspective projection collapses depth along optical rays, an object floating several centimeters above a tabletop projects to the identical 2D contour as a contacting object when scaled proportionally along its viewing ray. Open-loop single-pass synthesis therefore produces layouts that either misalign in image space or interpenetrate, topple, and drift under gravity.

To resolve projective and mechanical ambiguity jointly, the reconstruction agent $\mathcal{A}$ iterates through a two-stage verification loop inside a headless rendering and physics sandbox. The agent first defines a semantic object inventory with persistent identifiers and mobility roles, keeping movable foreground objects dynamic while locking supporting tables, floors, and walls as static colliders. In Stage~1 (\emph{visual mask-difference critique}), the agent synthesizes a procedural 3D script specifying manifold visual meshes, physically based materials, lighting, and camera parameters. Rendered per-object silhouettes form an aggregate foreground mask $M_{\mathrm{rend}}$, which is evaluated against the reference mask $M_{\mathrm{ref}}$ of $I_0$ via foreground IoU, a 72-bin radial centroid-to-contour profile, and Sobel interior edge overlap. A three-panel diagnostic overlay isolating reference-only deficits and render-only excess guides iterative updates to focal length, object dimensions, and 3D poses.

Visual alignment alone does not guarantee static support. In Stage~2 (\emph{zero-force physical equilibrium preflight}), the candidate world compiles into the dynamical solver with zero external actuation ($\mathbf{u}_t = \mathbf{0}$) and integrates under gravity $\mathbf{g}$ over $T_{\mathrm{eq}} = 1.0\,\mathrm{s}$. Acceptance requires bounded initial penetration, scale-calibrated translational or nodal rest, bounded terminal speeds, and numerical solver health across every dynamic object $i$:
\begin{equation}
\delta_{\max}(0) \le \varepsilon_c, \qquad
\max_{t \in [0, T_{\mathrm{eq}}]} \|\mathbf{x}_i(t) - \mathbf{x}_i(0)\|_2 \le \varepsilon_{\mathrm{eq},i}, \qquad
\|\dot{\mathbf{x}}_i(T_{\mathrm{eq}})\|_2 \le \varepsilon_{v,i}, \qquad
\text{finite}(\mathbf{s}_t) \;\land\; N_{\mathrm{warn}} = 0,
\label{eq:equilibrium_gate}
\end{equation}
alongside rigid-body orientation drift $\Delta\theta_i \le 5^\circ$ and terminal angular speed $\|\boldsymbol{\omega}_i(T_{\mathrm{eq}})\|_2 \le 0.1\,\mathrm{rad/s}$, where $\delta_{\max}(0)$ is maximum initial contact penetration and $\varepsilon_c$, $\varepsilon_{\mathrm{eq},i}$, and $\varepsilon_{v,i}$ denote feature-scaled tolerances (Table~\ref{tab:verification_thresholds}). When Stage~2 detects collider overlap, role mismatch, or gravitational drop, structured contact and drift traces feed back to the agent for geometric revision; on multi-body deformable stacks with occluded support clearances, damped quasi-static settling relaxes residual contact gaps before re-verifying Equation~\eqref{eq:equilibrium_gate} so that every exported world $\mathcal{W}_0$ begins at mechanical rest.

\subsection{Unified Multi-Regime Elastodynamics and Surface Perturbations}
\label{sec:method_physics}

Once verified, $\mathcal{W}_0$ maintains an explicit generalized state $\mathbf{s}_t = (\mathbf{q}_t, \mathbf{v}_t)$ independent of any neural video backbone and advances over discrete substeps $\Delta t$ via $\mathbf{s}_{t+1} = \mathcal{S}(\mathbf{s}_t, \mathbf{u}_t; \mathcal{W}_0)$. Our declarative schema compiles rigid, articulated, thin-shell, and volumetric deformable objects into a shared contact-coupled formulation \cite{todorov2012mujoco,macklin2016xpbd}.

\paragraph{Multi-Regime Constitutive Formulation.}
Rigid bodies evolve on $\mathrm{SE}(3)$ with six velocity coordinates per free body and convex primitive or decomposed colliders subject to elliptic frictional contact cones. Flexible 1D structures such as coiled cables are modeled as capsule chains linked by spherical joints that enforce segment inextensibility while transmitting axial tension and surface friction. Continuum deformables are discretized on a rest finite-element mesh $\mathcal{G} = (\mathcal{V}, \mathcal{E})$ with nodal positions $\mathbf{X}(t) = [\mathbf{x}_1(t), \dots, \mathbf{x}_V(t)]^\top \in \R^{V \times 3}$ and lumped masses $m_a > 0$. In generalized velocity coordinates $\mathbf{v}$, the coupled equations of motion take the form:
\begin{equation}
\mathbf{M}(\mathbf{q})\dot{\mathbf{v}} + \mathbf{c}(\mathbf{q}, \mathbf{v}) + \mathbf{f}_{\mathrm{int}}(\mathbf{q}, \mathbf{v}) = \mathbf{f}_{\mathrm{grav}}(\mathbf{q}) + \mathbf{J}_c(\mathbf{q})^\top \boldsymbol{\lambda}_c + \mathbf{f}_{\mathrm{ext}}(t),
\label{eq:elastodynamics}
\end{equation}
where $\mathbf{M}(\mathbf{q})$ combines rigid spatial inertias and deformable nodal masses, $\mathbf{c}(\mathbf{q}, \mathbf{v})$ collects Coriolis, centrifugal, and gyroscopic bias terms, $\mathbf{f}_{\mathrm{int}}(\mathbf{q}, \mathbf{v})$ represents passive elastic restoring and viscous damping generalized forces, and $\mathbf{J}_c(\mathbf{q})^\top \boldsymbol{\lambda}_c$ resolves unilateral normal and Coulomb frictional contacts across rigid-rigid, deformable-rigid, and deformable-deformable pairs via nodal contact spheres of radius $r_c$. Triangle meshes ($|\mathcal{E}_e| = 3$, $d = 2$) model woven and knitted textiles through coupled in-plane membrane stretching and cross-edge bending elasticity, complemented by compliant XPBD shell projection in Appendix~\ref{app:formulation}. Tetrahedral meshes ($|\mathcal{E}_e| = 4$, $d = 3$) model compressible foams and cushions via three-dimensional continuum elasticity parameterized by Young's modulus $E > 0$, Poisson's ratio $\nu \in [0, 0.5)$, and viscous damping. Implicit-velocity time integration stabilizes stiff membrane and volumetric modes while coupling deformable contact tractions bidirectionally to dynamic rigid bodies.

\paragraph{Calibrated Perturbations, State Continuation, and Counterfactuals.}
External actions $\mathbf{u}_t$ enter Equation~\eqref{eq:elastodynamics} either as localized surface and material loads or as impulsive and kinematic boundary conditions. For a 2D stroke drawn on $I_0$ from $\mathbf{u}_{\mathrm{start}}$ to $\mathbf{u}_{\mathrm{end}}$ with normalized screen length $\ell = \|\mathbf{u}_{\mathrm{end}} - \mathbf{u}_{\mathrm{start}}\|_2$, camera ray-casting locates the 3D surface anchor $\mathbf{x}_a(t_0)$ on rigid body $o_k$ and projects the stroke onto the camera-parallel plane along unit direction $\hat{\mathbf{d}} \in \R^3$. Weight-proportional scaling with $g = 9.81\,\mathrm{m/s^2}$ yields a world-frame force and center-of-mass torque over active window $w(t) \in [0, 1]$:
\begin{equation}
\mathbf{F}(t) = \alpha\,\ell\,m_k\,g\,w(t)\,\hat{\mathbf{d}}, \qquad
\boldsymbol{\tau}(t) = \bigl(\mathbf{x}_a(t) - \mathbf{x}_k(t)\bigr) \times \mathbf{F}(t),
\label{eq:surface_wrench}
\end{equation}
where gain $\alpha > 0$ and lever arm $\mathbf{r}(t) = \mathbf{x}_a(t) - \mathbf{x}_k(t)$ generate off-center tipping, pitching, and yawing moments around $\mathbf{x}_k(t)$. On deformable shells and 3D solids, manipulation acts as a distributed material-patch traction $\mathbf{F}_{\mathrm{patch}} \in \R^3$ over node subset $\mathcal{P} \subset \mathcal{V}$ with normalized weights $\sum_{a \in \mathcal{P}} w_a = 1$ and smooth $\sin^2$ temporal modulation:
\begin{equation}
\mathbf{f}_{\mathrm{ext},a}(t) = w_a\,\mathbf{F}_{\mathrm{patch}}\,w(t), \qquad a \in \mathcal{P}, \qquad \sum_{a \in \mathcal{P}} w_a = 1,
\label{eq:material_patch}
\end{equation}
preventing artificial single-vertex stress singularities. Instantaneous impulses increment generalized velocities via $\Delta \mathbf{v} = \mathbf{M}(\mathbf{q})^{-1}\mathbf{J}_{\mathrm{gen}}$ at event time $t_b$, whereas prescribed boundary trajectories drive designated attachment links along quintic smoothstep profiles $s(\tau) = 10\tau^3 - 15\tau^4 + 6\tau^5$. Complete solver checkpoints $\boldsymbol{\xi}_t$ serialize generalized coordinates $\mathbf{q}_t$, velocities $\mathbf{v}_t$, warm-start contact states, and active perturbation schedules with bit-exact determinism, enabling exact trajectory replay, live mid-rollout continuation at intermediate timestamp $t_b$, and counterfactual branching from identical initial states $\mathbf{s}_0$.

\subsection{Deformable Surface Skinning and Trajectory-Conditioned Video Diffusion}
\label{sec:method_video}

To render physics rollouts $V_{\mathrm{phys}} = \mathcal{R}(\mathbf{s}_{0:T}; \mathcal{W}_0)$ without first-frame pose snapping, each rigid visual mesh preserves its initial body-to-visual offset $\mathbf{T}_{\mathrm{rel},i} = \mathbf{T}_{\mathrm{body},i}(0)^{-1}\mathbf{T}_{\mathrm{vis},i}(0) \in \mathrm{SE}(3)$ and updates as $\mathbf{T}_{\mathrm{vis},i}(t) = \mathbf{T}_{\mathrm{body},i}(t)\mathbf{T}_{\mathrm{rel},i}$. Fine deformable visual meshes resolving hemmed borders, knitted ribs, or upholstery seams beyond the finite-element grid deform via precompiled linear-blend and normal-offset skinning. Each visual vertex $\mathbf{p}_r$ binds to $K$ simulation nodes $\{i_{r,j}\}_{j=1}^K$ with partition-of-unity weights $\sum_{j=1}^K w_{r,j} = 1$---spanning direct nodal ($K=1$), shell barycentric ($K=3$), and tetrahedral or trilinear cell ($K \in \{4, 8\}$) bindings---plus a signed thickness offset $d_r$ along its carrier triangle $(a_r, b_r, c_r)$:
\begin{equation}
\mathbf{p}_r(t) = \sum_{j=1}^{K} w_{r,j}\,\mathbf{x}_{i_{r,j}}(t) + d_r\,\hat{\mathbf{n}}_{r}(t), \qquad
\hat{\mathbf{n}}_{r}(t) = \frac{\bigl(\mathbf{x}_{b_r}(t) - \mathbf{x}_{a_r}(t)\bigr) \times \bigl(\mathbf{x}_{c_r}(t) - \mathbf{x}_{a_r}(t)\bigr)}{\bigl\|\bigl(\mathbf{x}_{b_r}(t) - \mathbf{x}_{a_r}(t)\bigr) \times \bigl(\mathbf{x}_{c_r}(t) - \mathbf{x}_{a_r}(t)\bigr)\bigr\|_2}.
\label{eq:skinning}
\end{equation}

A pretrained video generator $\mathcal{G}$ then synthesizes $\widehat{V} = \mathcal{G}\bigl(I_0, \Phi(V_{\mathrm{phys}}), p_{\mathrm{vid}}\bigr)$ anchored to $I_0$. Within structural diffusion adapters spanning Wan2.1-VACE 1.3B and Mixture-of-Experts Wan2.2-VACE-Fun A14B \cite{wan2025wan21,jiang2025vace,videoxfun2025}, we study two pixel-aligned control operators $\Phi$: sparse Canny contours $\Phi_{\mathrm{Canny}}(V_{\mathrm{phys}})$, which constrain silhouettes while leaving interior texture open to reference-guided synthesis, and dense shaded renders $\Phi_{\mathrm{RGB}}(V_{\mathrm{phys}})$, which provide depth-ordered occlusion and surface orientation cues. On rigid scenes, we additionally evaluate dual-reference audio-video synthesis with MiniMax-H3 Ref2VA \cite{minimax2025h3}, where $\Phi_{\mathrm{Ref}}(V_{\mathrm{phys}}) = V_{\mathrm{phys}}$, alongside unconditioned direct image-to-video baselines prompted from $I_0$ without simulated trajectories.

\FloatBarrier
\section{Experiments}

Our experiments evaluate three questions: whether closed-loop equilibrium verification is necessary for single-image 3D world recovery (\S\ref{sec:exp_loop}), how explicit multi-regime continuum mechanics compares against rigid-body approximations (\S\ref{sec:exp_mechanics}), and how simulated 3D rollouts govern trajectory fidelity and appearance preservation in video diffusion (\S\ref{sec:exp_video}).

\subsection{Closed-Loop Agentic Reconstruction vs.\ Single-Pass Open-Loop Synthesis}
\label{sec:exp_loop}

\begin{figure}[!t]
\centering
\includegraphics[width=\textwidth]{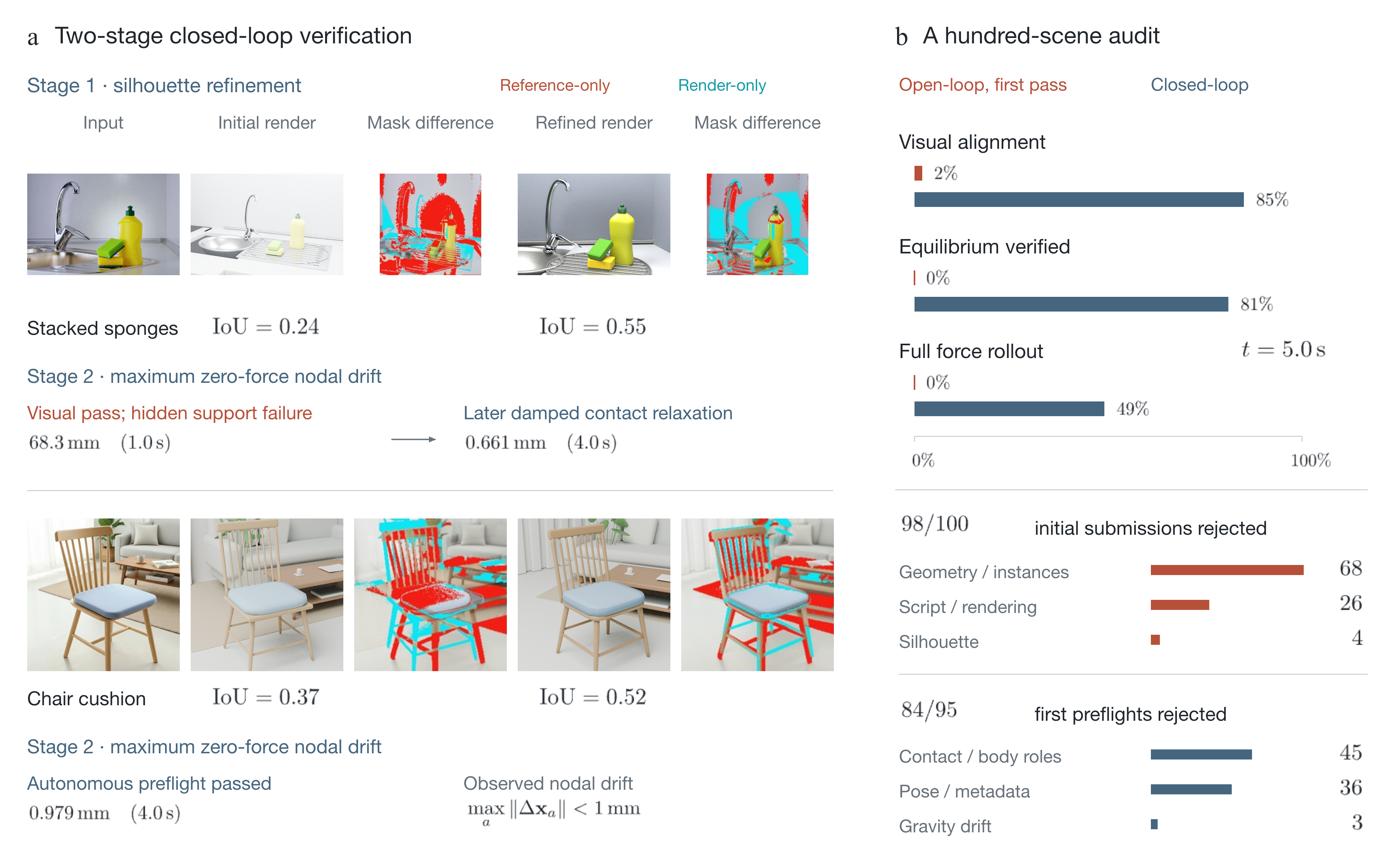}
\caption{\ClosedLoopCaption}
\label{fig:closed_loop}
\end{figure}

\begin{table}[!t]
\centering
\caption{\textbf{Controlled evaluation of single-pass open-loop versus two-stage closed-loop reconstruction.} Audited across the 100-scene single-image rigid tabletop benchmark and three single-image deformable and hybrid finite-element reconstructions.}
\label{tab:closed_loop_audit}
\footnotesize
\setlength{\tabcolsep}{4pt}
\begin{tabularx}{\textwidth}{l c c Y}
\toprule
\textbf{Benchmark \& Verification Metric} & \textbf{Open-Loop (Pass 1)} & \textbf{Closed-Loop} & \textbf{Dominant Failure Modes Diagnosed \& Repaired} \\
\midrule
\multicolumn{4}{@{}l}{\emph{100-Scene Single-Image Tabletop Benchmark ($N = 100$)}} \\
Stage-1 visual alignment pass & $2 / 100$ ($2.0\%$) & $85 / 100$ ($85.0\%$) & Pass-1 rejections: $68/100$ mesh topology or instance-binding errors, $26/100$ script/render exceptions, $4/100$ silhouette mismatches. \\
Stage-1 + Stage-2 equilibrium verified & $0 / 100$ ($0.0\%$) & $81 / 100$ ($81.0\%$) & First physics preflight ($95$ visually passing candidates): $45/95$ collider overlap or role errors, $36/95$ pose/inertia metadata errors, $3/95$ zero-force drift ($11/95$ pass on first check). \\
Full $5.0\,\mathrm{s}$ surface-force rollout verified & $0 / 100$ ($0.0\%$) & $49 / 100$ ($49.0\%$) & Verified packages converge in median $7$ scene submissions (mean $9.21$) and median $17$ tool calls (mean $18.72$). \\
\midrule
\multicolumn{4}{@{}l}{\emph{Single-Image Deformable \& Hybrid Finite-Element Reconstructions ($N = 3$: Bath Towel, Seat Cushion \& Chair, Kitchen Sponges \& Bottle)}} \\
Stage-1 visual + contact-clearance pass & $0 / 3$ ($0.0\%$) & $3 / 3$ ($100.0\%$) & Iterative camera/pose alignment reaches silhouette IoU $0.52\text{--}0.70$ with zero initial mesh interpenetration across all three scenes. \\
Autonomous $<1\,\mathrm{mm}$ nodal drift pass & $0 / 3$ ($0.0\%$) & $2 / 3$ ($66.7\%$) & Bath towel ($0.48\,\mathrm{mm}$ peak node drift) and chair cushion ($0.98\,\mathrm{mm}$ peak node drift) pass autonomously; Stage-2 catches a $68.3\,\mathrm{mm}$ nodal drop across an occluded support gap on the inclined sponge stack prior to damped relaxation. \\
\bottomrule
\end{tabularx}
\end{table}

\begin{figure}[!t]
\centering
\includegraphics[width=0.95\textwidth]{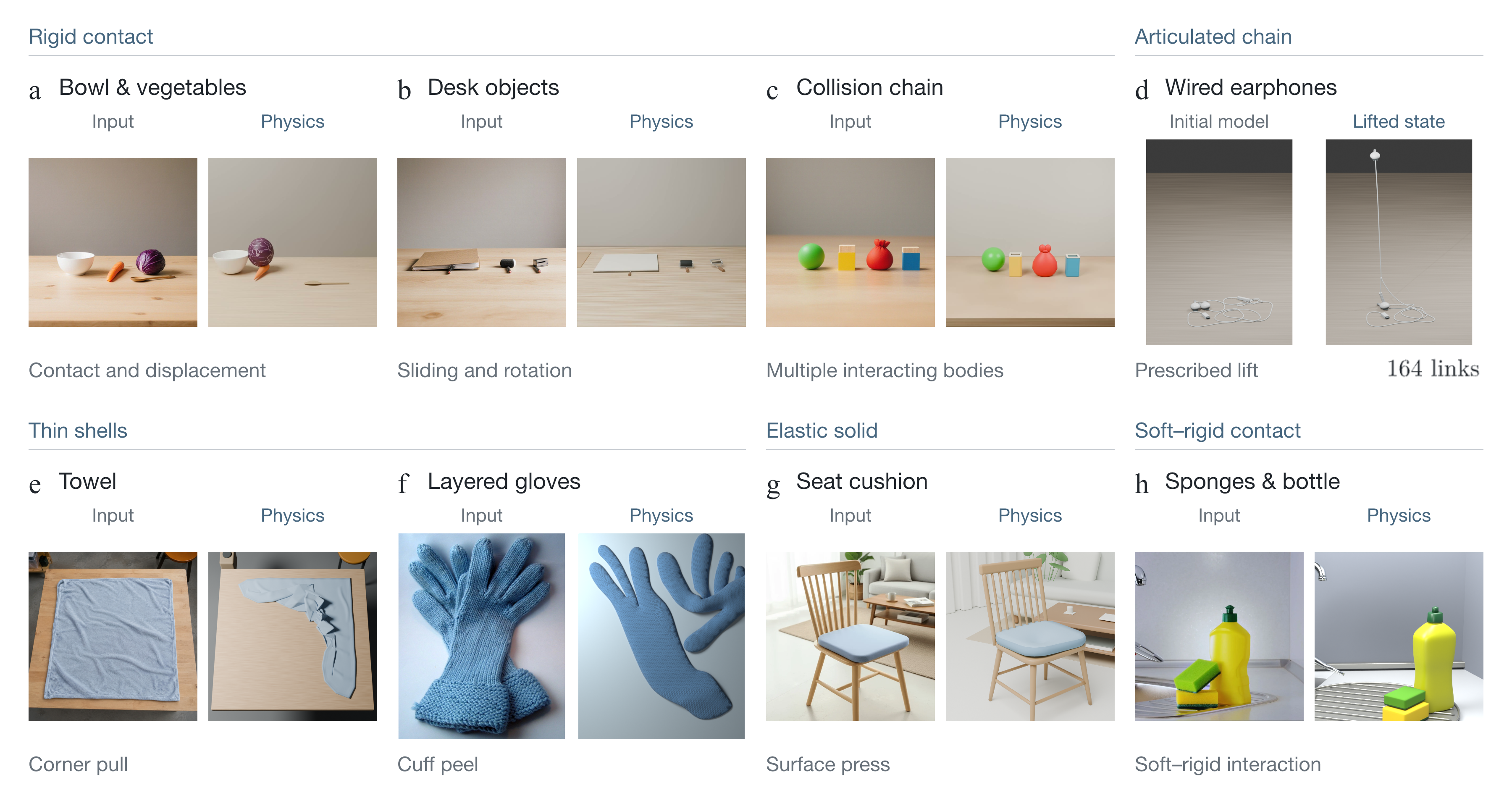}
\caption{\GalleryCaption}
\label{fig:gallery}
\end{figure}

We audit complete agent trajectories across our 100-scene single-image tabletop benchmark---covering household vessels, layered stationery, produce, lamps, footwear, and containers---together with three single-image deformable and hybrid reconstructions: a microfiber bath towel, an upholstered seat cushion on a wooden spindle-back chair, and stacked kitchen sponges beside a detergent bottle. Table~\ref{tab:closed_loop_audit} and Figure~\ref{fig:closed_loop} compare single-pass open-loop synthesis, evaluated on the agent's initial scene submission, against full two-stage closed-loop execution.

On the 100-scene benchmark, single-pass open-loop synthesis passes visual alignment on $2\%$ of inputs and achieves $0\%$ combined visual-and-physical verification. Two sequential failure barriers account for this result (Figure~\ref{fig:closed_loop}b). First, $98$ of $100$ initial scene scripts fail Stage~1 through non-manifold or misbound procedural meshes ($68\%$), script or renderer exceptions ($26\%$), or silhouette misalignment ($4\%$). Second, after iterative visual critique brings $95$ scenes across the Stage-1 visual threshold, $84$ of those $95$ visually aligned candidates ($88.4\%$) still fail their first Stage-2 physics preflight due to support-plane interpenetration or static/dynamic role errors ($45/95$), invalid quaternion or inertia specifications ($36/95$), or gravitational settling drift ($3/95$). Returning structured execution diagnostics resolves both barriers, recovering $85\%$ visually aligned scenes, $81\%$ equilibrium-verified world packages in a median of $7$ scene submissions and $17$ tool calls, and $49\%$ verified $5.0\,\mathrm{s}$ interactive surface-force rollouts.

The same two-stage verification governs single-image deformable and hybrid finite-element reconstruction (Figure~\ref{fig:closed_loop}a and Figure~\ref{fig:gallery}). Across all three scenes, closed-loop visual critique raises foreground IoU from $0.24\text{--}0.37$ to $0.52\text{--}0.70$ ($0.60\text{--}0.66$ composite visual score) with zero initial mesh interpenetration. Both the $121$-node bath towel and the $243$-node tetrahedral seat cushion satisfy the Stage-2 equilibrium gate autonomously within a single agent run, exhibiting $0.48\,\mathrm{mm}$ and $0.98\,\mathrm{mm}$ peak nodal drift. On the inclined three-body kitchen sponge stack, Stage~1 reaches clean 2D alignment (IoU $0.55$, silhouette structure $0.66$) with zero overlap, yet monocular occlusion hides a vertical gap beneath the tilted upper sponge. Stage~2 catches the resulting $68.3\,\mathrm{mm}$ peak nodal drop during zero-force preflight to prevent false visual-only acceptance, and damped contact relaxation subsequently establishes full three-body equilibrium ($0.66\,\mathrm{mm}$ peak nodal drift and $\le 0.57\,\mathrm{mm}$ CoM drift over $4.0\,\mathrm{s}$).

\subsection{Multi-Regime Stateful Mechanics, Counterfactuals, and Constitutive Ablation}
\label{sec:exp_mechanics}

\begin{figure}[!t]
\centering
\includegraphics[width=\textwidth]{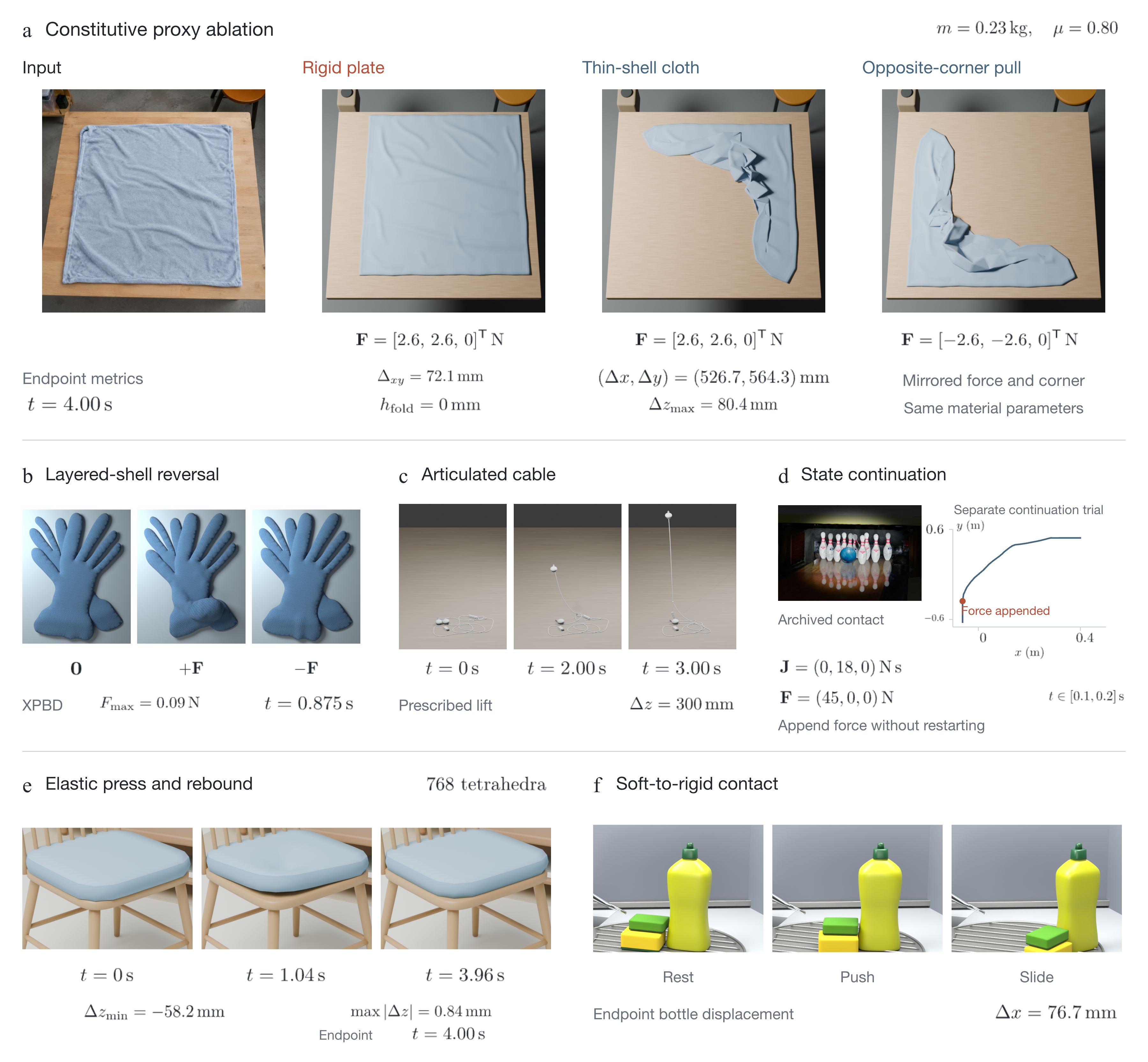}
\caption{\MechanicsCaption}
\label{fig:mechanics}
\end{figure}

\begin{table}[!t]
\centering
\caption{\textbf{Quantitative evaluation across seven constitutive and contact regimes.} Every scene is evaluated from a saved equilibrium state $\mathbf{s}_0$ under zero-force settling (terminal center-of-mass drift reported; peak nodal drift is $< 1.0\,\mathrm{mm}$ across all deformable cases), forward actuation, counterfactual reversal or live continuation, deterministic replay ($0.000\,\mathrm{mm}$ error across all cases), and half-timestep ($\Delta t/2$) step-size sensitivity.}
\label{tab:constitutive_mechanics}
\footnotesize
\setlength{\tabcolsep}{3.5pt}
\begin{tabularx}{\textwidth}{l l c Y}
\toprule
\textbf{Scene \& Regime} & \textbf{Discretization} & \textbf{Zero CoM Drift} & \textbf{Perturbation Protocol, Counterfactual / Ablation Contrast \& $\Delta t/2$ Sensitivity} \\
\midrule
Ten-pin bowling \newline (\emph{Rigid cascade}) & $11$ rigid bodies \newline ($\Delta t = 0.17\text{--}0.25\,\mathrm{ms}$) & $0.00\,\mathrm{mm}$ & $235.6\,\mathrm{N}$ horizontal launch topples $8/10$ pins; live mid-rollout continuation alters pin cascade to $9/10$ pins ($0.614\,\mathrm{m}$ divergence; $0.53\,\mathrm{mm}$ pre-impact $\Delta t/2$ diff). \\
Dual-pen holder \newline (\emph{Container clearance}) & $3$ rigid bodies \newline ($2.40\,\mathrm{mm}$ gap) & $\le 0.15\,\mathrm{mm}$ & Axial pulses ($0.39\,\mathrm{N}$, $0.59\,\mathrm{N}$) lift target pen $220\text{--}662\,\mathrm{mm}$ out of cup while unactuated neighbor shifts $< 10^{-15}\,\mathrm{m}$ ($0.05\,\mathrm{mm}$ airborne $\Delta t/2$ diff). \\
Wired earphones \newline (\emph{Articulated chain}) & $164$ capsule links \newline + $5$ terminal bodies & $\le 2.83\,\mathrm{mm}$ & Prescribed $+300\,\mathrm{mm}$ left-bud lift raises splitter $121.8\,\mathrm{mm}$ and controller $95.0\,\mathrm{mm}$ while frictional slack holds right bud to $1.26\,\mathrm{mm}$ ($1.42\,\mathrm{mm}$ RMS $\Delta t/2$ diff). \\
Bath towel \newline (\emph{Rigid-plate ablation}) & $1$ rigid plate \newline ($0.23\,\mathrm{kg}$, $\mu=0.80$) & $0.11\,\mathrm{mm}$ & Identical $[2.6, 2.6, 0]\,\mathrm{N}$ corner tug ($\sin^2$, $0.20\text{--}0.36\,\mathrm{s}$) yields $51.2\,\mathrm{mm}$ X, $50.8\,\mathrm{mm}$ Y flat slab shift ($72.1\,\mathrm{mm}$ norm, $0.0\,\mathrm{mm}$ fold; $0.11\,\mathrm{mm}$ $\Delta t/2$ RMS). \\
Bath towel \newline (\emph{Thin-shell cloth}) & $121$ nodes, $200$ tris \newline ($0.23\,\mathrm{kg}$, $\mu=0.80$) & $0.15\,\mathrm{mm}$ & Same $[2.6, 2.6, 0]\,\mathrm{N}$ tug pulls corner $527\,\mathrm{mm}$ X, $564\,\mathrm{mm}$ Y ($787\,\mathrm{mm}$ peak 3D) with $80\text{--}110\,\mathrm{mm}$ pleats; far corner moves $0.23\,\mathrm{mm}$ ($20.7\,\mathrm{mm}$ $\Delta t/2$ RMS). \\
Knitted wool gloves \newline (\emph{Layered shells}) & $2 \times 99$ nodes, $320$ tris \newline ($0.042\,\mathrm{kg}$ each) & $\le 0.24\,\mathrm{mm}$ & Oblique cuff traction ($[0.11, 0.025, 0.175]\,\mathrm{N}$, $0.25\text{--}1.05\,\mathrm{s}$) peels upper glove $386\,\mathrm{mm}$ peak while lower moves $0.67\,\mathrm{mm}$; $-\mathbf{F}(t)$ shifts $1.66\,\mathrm{mm}$ ($1.22\,\mathrm{mm}$ $\Delta t/2$ RMS). \\
Seat cushion \& chair \newline (\emph{3D solid + rigid}) & $243$ nodes, $768$ tets \newline + $4.5\,\mathrm{kg}$ chair & $\le 0.41\,\mathrm{mm}$ & Downward-oblique press ($[3.5, 4.0, -32.0]\,\mathrm{N}$, $0.25\text{--}1.55\,\mathrm{s}$) indents foam $-50.7\,\mathrm{mm}$ at $t=0.90\,\mathrm{s}$ ($-58.2\,\mathrm{mm}$ peak) and rebounds to $0.84\,\mathrm{mm}$ Z residual ($<0.7\,\mathrm{mm}$ $\Delta t/2$ CoM diff). \\
Sponges \& bottle \newline (\emph{3D hybrid contact}) & $2 \times 36$ nodes, $144$ tets \newline + $0.33\,\mathrm{kg}$ bottle & $\le 0.57\,\mathrm{mm}$ & Lateral squeeze ($[0.82, 0.04, 0]\,\mathrm{N}$ + $[0.52, 0.03, 0]\,\mathrm{N}$, $0.25\text{--}0.95\,\mathrm{s}$) compresses sponges against bottle, sliding bottle $+76.7\,\mathrm{mm}$ upright on wet steel ($\mu=0.18$; $0.3\,\mathrm{mm}$ $\Delta t/2$ X diff). \\
\bottomrule
\end{tabularx}
\end{table}

Figure~\ref{fig:mechanics} and Table~\ref{tab:constitutive_mechanics} analyze how constitutive laws govern contact isolation, deformability, and counterfactual interventions across rigid, articulated, thin-shell, and volumetric hybrid worlds.

\paragraph{Controlled Constitutive Proxy Ablation on Thin-Shell Cloth.}
To test whether thin flexible objects can be approximated by rigid collision proxies, we hold visual geometry, total mass ($0.23\,\mathrm{kg}$), tabletop Coulomb friction ($\mu = 0.80$), support geometry, and the corner force pulse ($[2.6, 2.6, 0]\,\mathrm{N}$ over $t \in [0.20, 0.36]\,\mathrm{s}$) fixed on the microfiber bath towel while replacing the $121$-node cloth shell with a matched rigid plate (Figure~\ref{fig:mechanics}a). Enforcing rigid-body kinematics across the $0.76 \times 0.80\,\mathrm{m}$ footprint requires overcoming normal friction across the entire sheet simultaneously, limiting motion to a $5.1\,\mathrm{cm}$ per-axis ($7.2\,\mathrm{cm}$ planar norm) slab shift with zero out-of-plane buckling. By contrast, the co-dimensional cloth shell decouples local membrane compliance from distant frictional anchorage: the actuated corner travels more than $10\times$ farther ($> 52\,\mathrm{cm}$ along both tabletop axes, $78.7\,\mathrm{cm}$ peak 3D displacement) and buckles into $8\text{--}11\,\mathrm{cm}$ diagonal pleats while the far corner stays pinned ($< 0.3\,\mathrm{mm}$ motion). Applying the mirrored counterfactual tug ($[-2.6, -2.6, 0]\,\mathrm{N}$) to the opposite corner patch from the same initial state $\mathbf{s}_0$ generates a symmetric fold along the orthogonal diagonal.

\paragraph{Layered Shell Disocclusion and High-DoF Articulated Tension.}
Explicit contact and joint constraints likewise isolate actuated branches from passive neighbors in multi-body arrangements (Figure~\ref{fig:mechanics}b--d). On the stacked knitted wool gloves ($2 \times 99$ shell nodes, with an $831$-vertex XPBD shell verification in Appendix~\ref{app:formulation}), upward-oblique cuff traction peels the upper glove $> 38\,\mathrm{cm}$ across the lower shell while tabletop friction holds the disoccluded lower glove below $0.7\,\mathrm{mm}$ drift. Reversing the traction vector ($-\mathbf{F}(t)$) presses the cuff into unilateral support, suppressing displacement by two orders of magnitude ($1.66\,\mathrm{mm}$). Along the $164$-segment wired earphone assembly, a prescribed $+30\,\mathrm{cm}$ lift on the left earbud transmits catenary tension to raise the inline controller and Y-splitter by $9.5\text{--}12.2\,\mathrm{cm}$, whereas slack loops beyond the splitter isolate the right earbud ($1.3\,\mathrm{mm}$ motion). In rigid ensembles, the same checkpointed solver maintains sub-femtometer neighbor isolation across a $2.4\,\mathrm{mm}$ pen-holder gap and supports live mid-rollout force or impulse continuation in ten-pin bowling without resetting $\mathbf{s}_0$.

\paragraph{Volumetric Elastic Recovery and Soft-to-Rigid Momentum Transfer.}
Extending from 2D shells to 3D tetrahedral elastodynamics ($d = 3$) captures through-thickness compression and compliant momentum transfer to movable rigid bodies (Figure~\ref{fig:mechanics}e--f). Under a downward-oblique load peaking at $32\,\mathrm{N}$ vertical force, the $768$-tetrahedron seat cushion reaches $-5.8\,\mathrm{cm}$ peak nodal indentation ($-5.1\,\mathrm{cm}$ at $t = 0.90\,\mathrm{s}$) atop a dynamic $4.5\,\mathrm{kg}$ wooden chair and recovers elastically upon unloading to within $0.84\,\mathrm{mm}$ of its initial height. In the hybrid kitchen sink scene, lateral traction on two stacked $36$-node foam sponges ($30\,\mathrm{g}$ each, $E = 4.0\,\mathrm{kPa}$) compresses the compliant foam interface against a $0.33\,\mathrm{kg}$ rigid detergent bottle and transfers horizontal impulse across the contact patch, sliding the bottle $+76.7\,\mathrm{mm}$ upright along the wet steel drainboard ($\mu = 0.18$) with $0.3\,\mathrm{mm}$ X-axis bottle travel sensitivity under $\Delta t/2$ refinement.

\subsection{Physics-Grounded Video Synthesis vs.\ Direct Image-to-Video Generation}
\label{sec:exp_video}

\begin{table}[!t]
\centering
\caption{\textbf{Quantitative comparison of physics-grounded video synthesis against unconditioned direct image-to-video baselines.} Evaluated across Rigid ($N=3$), Deformable \& Hybrid ($N=4$), All Primary ($N=7$), and Counterfactual opposite towel pull ($N=1$) subsets. Arrows indicate whether higher ($\uparrow$) or lower ($\downarrow$) values reflect better performance; $\cos\theta_{\mathrm{dir}}$ measures net motion-centroid directional alignment with the simulated trajectory.}
\label{tab:video_metrics}
\scriptsize
\setlength{\tabcolsep}{2.2pt}
\begin{tabular*}{\textwidth}{@{\extracolsep{\fill}}l l c c c c c c r r@{}}
\toprule
\textbf{Subset} & \textbf{Generator \& Control} & \textbf{Mask IoU $\uparrow$} & \textbf{Traj (\%) $\downarrow$} & \textbf{PSNR $\uparrow$} & \textbf{SSIM $\uparrow$} & \textbf{DINOv2 (Full/ROI) $\uparrow$} & \textbf{$\cos\theta_{\mathrm{dir}}$ $\uparrow$} & \textbf{VRAM} & \textbf{Time (s)} \\
\midrule
\multirow{7}{*}{\shortstack[l]{\textbf{Rigid}\\($N = 3$)}}
 & Direct I2V 1.3B (No Phys) & 0.1906 & 12.58 & \textbf{25.71} & 0.9351 & 0.8021 / 0.7862 & $+0.6398$ & 19.4\,GB & 74.5 \\
 & Direct I2V A14B (No Phys) & 0.2301 & 10.40 & 22.49 & 0.9520 & 0.7643 / 0.7210 & $-0.2053$ & 41.2\,GB & 354.4 \\
 & Wan2.1-VACE 1.3B (Canny) & 0.4556 & 3.84 & 10.21 & 0.6918 & 0.5912 / 0.6174 & $+0.8161$ & 19.4\,GB & 55.9 \\
 & Wan2.1-VACE 1.3B (RGB) & \textbf{0.8424} & \textbf{1.00} & 16.87 & 0.8676 & 0.6502 / 0.5696 & \textbf{$+0.9549$} & 19.4\,GB & \textbf{55.1} \\
 & Wan2.2-VACE A14B (Canny) & 0.6936 & 2.31 & 22.29 & 0.9286 & 0.6416 / 0.6526 & $+0.3080$ & 41.0\,GB & 291.4 \\
 & Wan2.2-VACE A14B (RGB) & 0.6352 & 3.42 & 23.46 & 0.9306 & 0.6898 / 0.6540 & $+0.5358$ & 41.0\,GB & 257.7 \\
 & MiniMax-H3 Ref2VA$^\dagger$ & 0.5141 & 4.05 & 24.68 & \textbf{0.9544} & \textbf{0.9368 / 0.9134} & $-0.0996$ & 35.1\,GB & 66.4 \\
\midrule
\multirow{6}{*}{\shortstack[l]{\textbf{Deform.}\\($N = 4$)}}
 & Direct I2V 1.3B (No Phys) & 0.1617 & 20.94 & \textbf{25.59} & 0.8596 & \textbf{0.9607 / 0.9567} & $+0.1631$ & 19.4\,GB & 77.7 \\
 & Direct I2V A14B (No Phys) & 0.2948 & 16.65 & 22.77 & \textbf{0.9002} & 0.8605 / 0.8437 & $-0.1986$ & 41.2\,GB & 314.2 \\
 & Wan2.1-VACE 1.3B (Canny) & 0.4910 & 4.84 & 10.80 & 0.5583 & 0.8388 / 0.8353 & $+0.7075$ & 19.4\,GB & 77.4 \\
 & Wan2.1-VACE 1.3B (RGB) & \textbf{0.8850} & \textbf{0.75} & 13.98 & 0.6689 & 0.6399 / 0.6260 & $+0.9212$ & 19.4\,GB & \textbf{77.2} \\
 & Wan2.2-VACE A14B (Canny) & 0.5659 & 3.80 & 18.41 & 0.7046 & 0.8365 / 0.8264 & \textbf{$+0.9293$} & 41.0\,GB & 296.8 \\
 & Wan2.2-VACE A14B (RGB) & 0.6002 & 2.69 & 17.89 & 0.6945 & 0.8202 / 0.8041 & $+0.6540$ & 41.0\,GB & 292.2 \\
\midrule
\multirow{6}{*}{\shortstack[l]{\textbf{Primary}\\($N = 7$)}}
 & Direct I2V 1.3B (No Phys) & 0.1741 & 17.35 & \textbf{25.64} & 0.8919 & \textbf{0.8927 / 0.8836} & $+0.3674$ & 19.4\,GB & 76.3 \\
 & Direct I2V A14B (No Phys) & 0.2670 & 13.97 & 22.65 & \textbf{0.9224} & 0.8193 / 0.7911 & $-0.2015$ & 41.2\,GB & 331.4 \\
 & Wan2.1-VACE 1.3B (Canny) & 0.4758 & 4.41 & 10.55 & 0.6155 & 0.7327 / 0.7419 & $+0.7540$ & 19.4\,GB & 68.2 \\
 & Wan2.1-VACE 1.3B (RGB) & \textbf{0.8667} & \textbf{0.86} & 15.22 & 0.7541 & 0.6443 / 0.6019 & \textbf{$+0.9356$} & 19.4\,GB & \textbf{67.7} \\
 & Wan2.2-VACE A14B (Canny) & 0.6206 & 3.16 & 20.07 & 0.8006 & 0.7530 / 0.7519 & $+0.6187$ & 41.0\,GB & 294.5 \\
 & Wan2.2-VACE A14B (RGB) & 0.6152 & 3.00 & 20.28 & 0.7957 & 0.7643 / 0.7398 & $+0.6033$ & 41.0\,GB & 277.4 \\
\midrule
\multirow{4}{*}{\shortstack[l]{\textbf{Opposite}\\(\emph{Towel})}}
 & Direct I2V 1.3B (No Phys) & 0.0043 & 34.30 & \textbf{30.15} & \textbf{0.9164} & \textbf{0.9900 / 0.9878} & $-0.7441$ & 19.4\,GB & 69.7 \\
 & Direct I2V A14B (No Phys) & 0.4444 & 18.51 & 25.49 & 0.9012 & 0.7890 / 0.7898 & $-0.1702$ & 41.2\,GB & 347.9 \\
 & Wan2.2-VACE A14B (Canny) & \textbf{0.5713} & 13.30 & 17.15 & 0.6813 & 0.5130 / 0.5214 & \textbf{$+0.9579$} & 41.0\,GB & 356.4 \\
 & Wan2.2-VACE A14B (RGB) & 0.5680 & \textbf{10.34} & 16.75 & 0.6355 & 0.5265 / 0.5305 & $+0.9083$ & 41.0\,GB & 337.7 \\
\bottomrule
\multicolumn{10}{@{}l@{}}{\scriptsize $^\dagger$Evaluated on the three rigid benchmark scenes at $768 \times 768$ ($124$ frames at $24\,\mathrm{fps}$, $8{\times}\text{H100}$ FSDP/Ulysses, VRAM per GPU) with $32\,\mathrm{kHz}$ stereo audio.}
\end{tabular*}
\end{table}

\begin{figure}[!t]
\centering
\includegraphics[width=\textwidth]{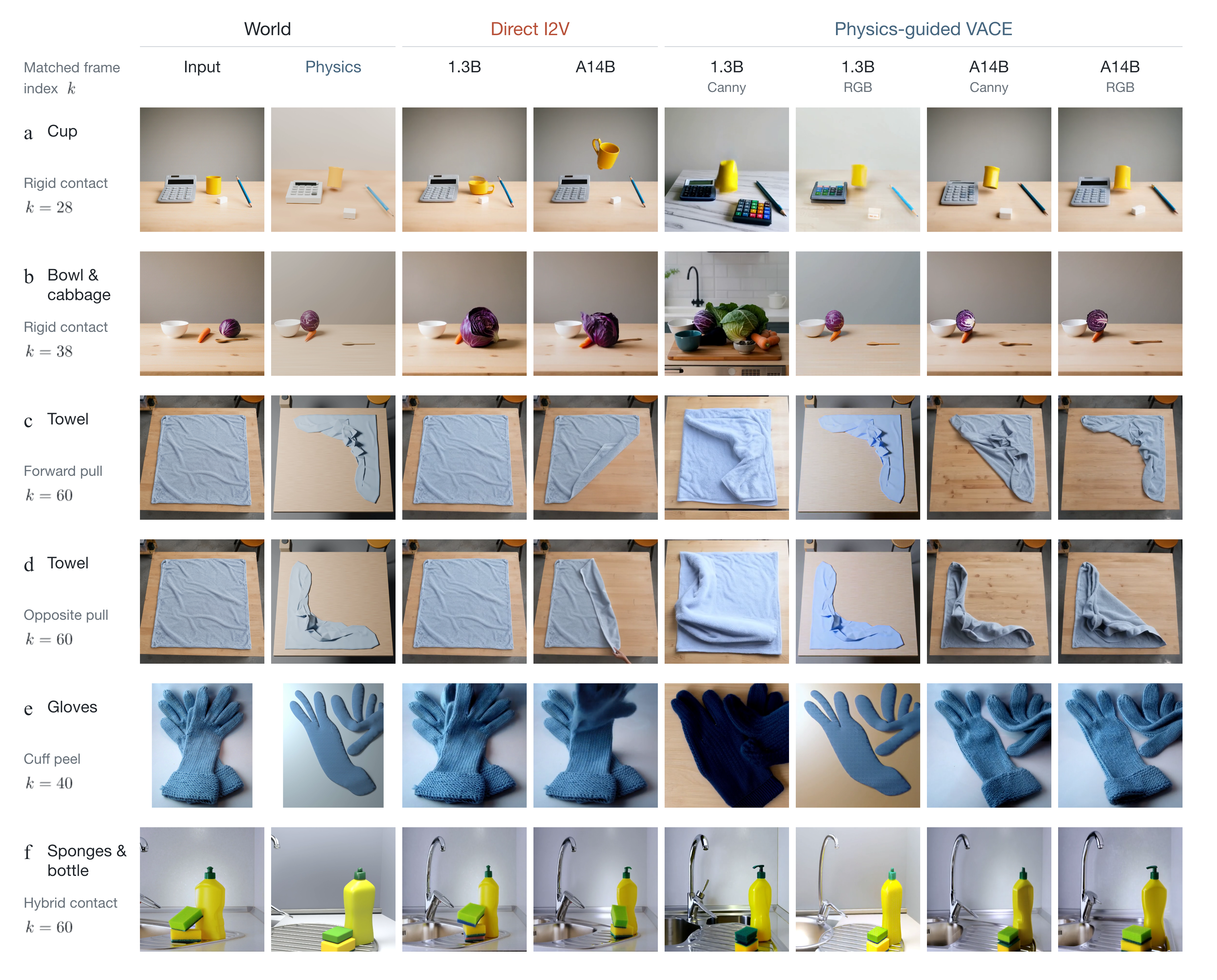}
\caption{\VideoComparisonCaption}
\label{fig:video_comparison}
\end{figure}

We evaluate generative video grounding across seven primary interactions---ceramic mug tumbling, kraft notebook sliding, vegetable-bowl collision, bath towel corner pulling, knitted wool glove peeling, sponge-to-bottle pushing, and seat cushion indentation---together with two counterfactual queries testing opposite-corner towel pulling and horizontal mug sliding ($53$ generated videos in total; Table~\ref{tab:video_metrics} and Appendix~\ref{app:per_scene_ledger}). Four trajectory-conditioned VACE variants spanning Wan2.1-VACE 1.3B and Wan2.2-VACE-Fun A14B under $\Phi_{\mathrm{Canny}}$ and $\Phi_{\mathrm{RGB}}$ ($512 \times 512$, $81$ frames, $30$ denoising steps) are compared against unconditioned Direct I2V 1.3B and A14B baselines anchored on the same input photograph $I_0$ and action prompts without $V_{\mathrm{phys}}$, alongside rigid-scene MiniMax-H3 Ref2VA rollouts. Performance is quantified via dynamic motion-mask IoU, normalized motion-centroid trajectory error (as image-width percentage), static-background PSNR and SSIM relative to $I_0$, full-frame and region-of-interest DINOv2 ViT-L/14-reg4 \cite{oquab2024dinov2} cosine similarity, and net motion-direction cosine alignment $\cos\theta_{\mathrm{dir}}$.

\paragraph{Failure of Unconditioned Direct I2V on Stateful and Counterfactual Actions.}
Table~\ref{tab:video_metrics} and Figure~\ref{fig:video_comparison} reveal a fundamental tension in unconditioned direct image-to-video synthesis between static frame copying and ungrounded motion hallucination. Direct I2V 1.3B attains high static PSNR ($> 25\,\mathrm{dB}$) and DINOv2 similarity ($0.89$) primarily by leaving deformable scenes nearly frozen: its dynamic motion-mask IoU falls below $0.03$ on both the bath towel and the seat cushion, yielding $> 20\%$ mean centroid trajectory error across deformable and hybrid scenes. Scaling to Direct I2V A14B increases pixel motion but still lacks 3D contact or force grounding: it levitates the ceramic mug mid-air, expands the red cabbage in place rather than rolling it into the bowl, and morphs glove fingers without peeling the upper shell. Under counterfactual cloth prompts, Direct I2V A14B folds the lower-right towel corner regardless of whether the prompt specifies pulling the lower-left or upper-right corner, producing negative directional alignment ($\cos\theta_{\mathrm{dir}} \approx -0.20$ across primary scenes and $-0.17$ on the opposite pull). Conditioning VACE on simulated 3D rollouts raises overall motion-mask IoU by $2.3{\times}\text{--}5.0{\times}$, cuts centroid trajectory error from $14\%\text{--}17\%$ to $0.9\%\text{--}4.4\%$, and restores $> +0.90$ directional cosine alignment across both forward and counterfactual cloth pulls.

\paragraph{Backbone Capacity and Conditioning Modality Trade-Off.}
Comparing $\Phi_{\mathrm{Canny}}$ and $\Phi_{\mathrm{RGB}}$ across the 1.3B and 14B VACE backbones clarifies how structural control density interacts with denoiser capacity. On the 1.3B backbone, sparse Canny contours ($\Phi_{\mathrm{Canny}}$) leave background illumination and interior appearance weakly constrained, lowering static-background fidelity to $\sim 10.6\,\mathrm{dB}$ PSNR with visible tabletop texture drift (Figure~\ref{fig:video_comparison}, column~5). Dense RGB renders ($\Phi_{\mathrm{RGB}}$) on 1.3B lock silhouettes tightly to the simulated trajectory (IoU $0.87$, trajectory error $< 1\%$), but cause the compact denoiser to copy flat synthetic mesh shading onto deformable textiles, reducing DINOv2 similarity to $0.64$ on deformable scenes. Scaling to the 14B Mixture-of-Experts backbone (Wan2.2-VACE-Fun A14B) eliminates both failure modes, improving static-background PSNR by $+5\text{--}10\,\mathrm{dB}$ over 1.3B while preserving microfiber weaves, knitted wool ribs, steel reflections, and upholstery creases ($> 0.82$ DINOv2 similarity on deformable and hybrid scenes) at $2.1\times$ higher GPU memory ($41.0\,\mathrm{GB}$ vs.\ $19.4\,\mathrm{GB}$). On rigid scenes, dual-reference MiniMax-H3 Ref2VA complements pixel-aligned VACE control by preserving $768^2$ photographic reflectance ($0.94$ DINOv2 similarity, $24.7\,\mathrm{dB}$ static PSNR) and synthesizing synchronized $32\,\mathrm{kHz}$ impact audio with looser pixel-level trajectory locking.

\section{Conclusion}

We presented TelePhysica, a closed-loop framework that reconstructs executable, multi-regime 3D physical worlds from a single photograph to drive stateful interaction and trajectory-grounded video synthesis. Pairing pixel-aligned visual critique with a zero-force gravitational equilibrium preflight raises verified single-image scene recovery from $0\%$ in open-loop synthesis to $81\%$ across 100 tabletop scenes, while unified rigid, articulated, thin-shell, and 3D tetrahedral elastodynamics captures localized cloth folding, elastic foam rebound, and soft-to-rigid momentum transfer for controllable video diffusion. Constitutive parameters such as mass density, Young's modulus, and friction coefficients are currently assigned from material priors rather than identified online from multi-step visual interaction feedback, defining a clear avenue for closed-loop physical system identification.

\clearpage
\appendix

\section{Mathematical Formulation, Simulation Conventions, and Verification Criteria}
\label{app:formulation}

\paragraph{Coordinate Frame, Units, and Pose Conventions.}
All scene scripts, solvers, and perturbation schedules share a right-handed Cartesian world frame with $+X$ right, $+Y$ forward in depth, and $+Z$ vertically upward against gravity $\mathbf{g} = (0, 0, -9.81)^\top\,\mathrm{m/s^2}$. Lengths are expressed in meters ($\mathrm{m}$), masses in kilograms ($\mathrm{kg}$), time in seconds ($\mathrm{s}$), forces in newtons ($\mathrm{N}$), moments in newton-meters ($\mathrm{N\cdot m}$), linear impulses in newton-seconds ($\mathrm{N\cdot s}$), and elastic moduli in pascals ($\mathrm{Pa}$). Rigid-body orientations use scalar-first unit quaternions $\mathbf{q} = (w, x, y, z) \in S^3$ ($\|\mathbf{q}\|_2 = 1$) mapped to orthogonal matrices $\mathbf{R}(\mathbf{q}) \in \mathrm{SO}(3)$ when composing spatial transforms $\mathbf{T} \in \mathrm{SE}(3)$.

\paragraph{Native Finite-Element Flex Compilation and Compliant XPBD Thin Shells.}
In our declarative schema, deformable objects specify persistent node identifiers, world-space rest coordinates $\bar{\mathbf{X}} \in \R^{V \times 3}$, optional prestressed initial coordinates $\mathbf{X}(0) \in \R^{V \times 3}$, homogeneous element connectivity $\mathcal{E}$ using triangles for $d = 2$ shells and tetrahedra for $d = 3$ solids, pinned anchor subsets, and constitutive parameters $(m, E, \nu, \eta, \mu, r_c)$. During native MuJoCo flexible-body compilation \cite{todorov2012mujoco}, world-space nodes map into the parent body frame via $\mathbf{x}_a^{\mathrm{loc}} = \mathbf{R}_0^\top(\mathbf{x}_a - \mathbf{p}_0)$, and lumped nodal masses distribute across vertex bodies carrying three translational degrees of freedom per free node. Implicit-velocity integration then advances membrane/bending ($d=2$) or volumetric continuum ($d=3$) elastic forces jointly with unilateral frictional contact constraints.

For an $831$-vertex high-resolution knitted wool glove shell ($499$ front-glove and $332$ rear-glove midsurface vertices, $0.035\,\mathrm{kg}$ per glove, $\Delta t = 1/480\,\mathrm{s}$), we additionally verify Extended Position-Based Dynamics (XPBD) \cite{muller2007pbd,macklin2016xpbd} with edge-stretch stiffness $k_{\mathrm{stretch}} = 120.0\,\mathrm{N/m}$, cross-edge bending stiffness $k_{\mathrm{bend}} = 0.8\,\mathrm{N/m}$, and compliance $\tilde{\alpha} = 1/(k\Delta t^2)$. Within each Gauss-Seidel substep iteration, Lagrange multiplier increments update nodal pairs $(a, b)$ along $\hat{\mathbf{n}}_{ab} = (\tilde{\mathbf{x}}_a - \tilde{\mathbf{x}}_b)/\|\tilde{\mathbf{x}}_a - \tilde{\mathbf{x}}_b\|_2$ according to:
\begin{equation}
\Delta\lambda_{ab} = -\frac{\|\tilde{\mathbf{x}}_a - \tilde{\mathbf{x}}_b\|_2 - L_{ab}^0 + \tilde{\alpha}\,\lambda_{ab}}{m_a^{-1} + m_b^{-1} + \tilde{\alpha}}, \qquad
\Delta\tilde{\mathbf{x}}_a = +m_a^{-1}\Delta\lambda_{ab}\,\hat{\mathbf{n}}_{ab}, \qquad
\Delta\tilde{\mathbf{x}}_b = -m_b^{-1}\Delta\lambda_{ab}\,\hat{\mathbf{n}}_{ab}.
\label{eq:xpbd_update}
\end{equation}
Over a $3.0\,\mathrm{s}$ zero-input settle, center-of-mass drift is $0.012\,\mathrm{mm}$ ($0.226\,\mathrm{mm}$ peak vertex drift). A localized $0.09\,\mathrm{N}$ front-cuff $\sin^2$ traction pulse ($t \in [0.35, 0.95]\,\mathrm{s}$) produces $75.99\,\mathrm{mm}$ peak vertex lift ($10.59\,\mathrm{mm}$ CoM shift), reversed traction $-\mathbf{F}(t)$ yields $3.87\,\mathrm{mm}$ peak vertex motion against tabletop friction ($\mu = 0.55$), deterministic replay error is $0.000\,\mathrm{mm}$, and half-step refinement ($\Delta t/2 = 1/960\,\mathrm{s}$) gives a $3.32\,\mathrm{mm}$ terminal vertex RMS difference ($4.37\%$ of peak).

\begin{table}[!htbp]
\centering
\caption{\textbf{Automated geometric and physical verification gates enforced across scene reconstruction and rollout evaluation.}}
\label{tab:verification_thresholds}
\footnotesize
\begin{tabularx}{\textwidth}{l l Y}
\toprule
\textbf{Verification Stage} & \textbf{Evaluated Quantity} & \textbf{Quantitative Acceptance Criterion} \\
\midrule
Role Inventory & Mobility role $\rho_i$ & Every scene object declares an explicit role; movable foreground bodies and deformables are assigned \texttt{dynamic}. \\
Mesh Topology & Manifoldness \& area & Zero non-manifold edges ($\le 2$ faces per edge) and zero degenerate triangles below scale-normalized area $10^{-10}\,\max(\mathrm{ptp}(\mathbf{V}), 1)^2$. \\
Stage-1 Visual Gate & Silhouette \& edge IoU & Combined occupancy, 72-bin radial contour profile, and Sobel interior edge score $\ge 0.45$ with foreground mask IoU $\ge 0.45$ and silhouette structure $\ge 0.60$. \\
Stage-2 Equilibrium & Initial contact \& $1.0\,\mathrm{s}$ settle & Pairwise initial overlap $\delta_{ij}(0) \le \max(0.1\,\mathrm{mm}, 0.01\min_{k \in \mathcal{D}_{ij}}\ell_k)$ ($\ell_k$: smallest declared primitive collider thickness of dynamic object $k$ in pair), drift $\le \varepsilon_{\mathrm{eq},i} = \max(1.0\,\mathrm{mm}, 0.02\,\ell_i)$, terminal speed $\le \varepsilon_{v,i}$ ($\max(0.01\,\mathrm{m/s}, 0.05\,\ell_i)$ rigid; $\max(0.01\,\mathrm{m/s}, 0.05\,\varepsilon_{\mathrm{eq},i})$ deformable), rigid rotation $\le 5^\circ$, angular speed $\le 0.1\,\mathrm{rad/s}$, and $0$ solver warnings. \\
Perturbation Rollout & Multi-trial execution & Schema-validated interventions execute across forward, negated counterfactual ($-\mathbf{F}$), replay, and $\Delta t/2$ trials with finite integration state, $0$ solver warnings, and logged force-time integrals $\int \mathbf{F}(t)\,\dd t$ and $\Delta t/2$ RMS sensitivity. \\
State Determinism & Checkpoint \& continuation & Full-state checkpoint replay, midpoint pause/restore continuation, and chunked stepping reproduce complete integration state vectors within $\le 10^{-12}$ maximum absolute error. \\
\bottomrule
\end{tabularx}
\end{table}

\section{Per-Scene Video Synthesis Quantitative Ledger}
\label{app:per_scene_ledger}

Table~\ref{tab:per_scene_video_ledger} reports the complete per-scene quantitative evaluation across all seven primary multi-regime scenes and two counterfactual action queries ($53$ evaluated video generations across Direct I2V 1.3B, Direct I2V A14B, Wan2.1-VACE 1.3B Canny/RGB, Wan2.2-VACE-Fun A14B Canny/RGB, and MiniMax-H3 Base Ref2VA). On the upholstered seat cushion scene, downward indentation is followed by elastic recovery to within $0.84\,\mathrm{mm}$ of the initial surface height at $t = 4.0\,\mathrm{s}$; start-to-end 2D centroid translation is therefore near zero by physical design, whereas framewise dynamic motion-mask IoU and trajectory error capture the transient press-and-rebound cycle.

\begin{table}[!htbp]
\centering
\caption{\textbf{Complete per-scene quantitative video generation ledger across rigid, thin-shell deformable, 3D hybrid, and counterfactual evaluations.}}
\label{tab:per_scene_video_ledger}
\fontsize{7.0}{8.4}\selectfont
\setlength{\tabcolsep}{2.6pt}
\setlength{\aboverulesep}{1.2pt}
\setlength{\belowrulesep}{1.5pt}
\renewcommand{\arraystretch}{0.88}
\begin{tabular*}{\textwidth}{@{\extracolsep{\fill}}l l c c c c c c r@{}}
\toprule
\textbf{Scene (Regime)} & \textbf{Model \& Conditioning} & \textbf{Mask IoU $\uparrow$} & \textbf{Traj Err (\%) $\downarrow$} & \textbf{PSNR (dB) $\uparrow$} & \textbf{SSIM $\uparrow$} & \textbf{DINOv2 (Full / ROI) $\uparrow$} & \textbf{$\cos\theta_{\mathrm{dir}}$ $\uparrow$} & \textbf{Time (s)} \\
\midrule
\multirow{7}{*}{\shortstack[l]{Yellow Ceramic Mug\\(\emph{Rigid tumble})}}
 & Direct I2V 1.3B & 0.1004 & 13.57 & 30.49 & 0.9660 & 0.9525 / 0.9292 & $+0.3515$ & 69.0 \\
 & Direct I2V A14B & 0.2070 & 8.99 & 24.26 & 0.9613 & 0.8562 / 0.8319 & $-0.5519$ & 376.1 \\
 & VACE 1.3B (Canny) & 0.4968 & 3.63 & 11.14 & 0.7200 & 0.8253 / 0.7989 & $+0.9079$ & 55.8 \\
 & VACE 1.3B (RGB) & 0.7125 & 2.07 & 12.04 & 0.8397 & 0.7480 / 0.6426 & $+0.9953$ & 56.0 \\
 & VACE A14B (Canny) & 0.6194 & 3.64 & 18.46 & 0.8747 & 0.8738 / 0.7668 & $+0.5897$ & 253.6 \\
 & VACE A14B (RGB) & 0.4334 & 7.64 & 18.90 & 0.8746 & 0.8367 / 0.7964 & $+0.9296$ & 245.4 \\
 & MiniMax-H3 Ref2VA & 0.1888 & 9.28 & 24.48 & 0.9662 & 0.9542 / 0.9373 & $-0.7232$ & 66.4 \\
\midrule
\multirow{7}{*}{\shortstack[l]{Kraft Notebook \& Pencils\\(\emph{Rigid slide})}}
 & Direct I2V 1.3B & 0.2268 & 8.59 & 23.11 & 0.9395 & 0.7480 / 0.7520 & $+0.6178$ & 74.5 \\
 & Direct I2V A14B & 0.2322 & 4.98 & 21.90 & 0.9477 & 0.6695 / 0.5921 & $-0.8377$ & 314.2 \\
 & VACE 1.3B (Canny) & 0.4919 & 1.18 & 8.43 & 0.6735 & 0.2945 / 0.3513 & $+0.5428$ & 55.6 \\
 & VACE 1.3B (RGB) & 0.8779 & 0.68 & 18.70 & 0.8222 & 0.5024 / 0.4667 & $+0.8693$ & 55.0 \\
 & VACE A14B (Canny) & 0.7091 & 1.98 & 21.04 & 0.9329 & 0.5220 / 0.6715 & $-0.6634$ & 301.1 \\
 & VACE A14B (RGB) & 0.7384 & 1.24 & 22.81 & 0.9427 & 0.7470 / 0.6826 & $-0.3003$ & 251.7 \\
 & MiniMax-H3 Ref2VA & 0.6060 & 2.03 & 24.47 & 0.9350 & 0.9443 / 0.8798 & $-0.5745$ & 66.3 \\
\midrule
\multirow{7}{*}{\shortstack[l]{Bowl \& Red Cabbage\\(\emph{Rigid collision})}}
 & Direct I2V 1.3B & 0.2446 & 15.58 & 23.52 & 0.8997 & 0.7057 / 0.6774 & $+0.9500$ & 79.9 \\
 & Direct I2V A14B & 0.2510 & 17.23 & 21.31 & 0.9471 & 0.7673 / 0.7389 & $+0.7736$ & 372.9 \\
 & VACE 1.3B (Canny) & 0.3782 & 6.71 & 11.05 & 0.6820 & 0.6539 / 0.7021 & $+0.9977$ & 56.3 \\
 & VACE 1.3B (RGB) & 0.9369 & 0.25 & 19.88 & 0.9409 & 0.7003 / 0.5996 & $+1.0000$ & 54.4 \\
 & VACE A14B (Canny) & 0.7524 & 1.32 & 27.37 & 0.9783 & 0.5290 / 0.5194 & $+0.9976$ & 319.5 \\
 & VACE A14B (RGB) & 0.7337 & 1.37 & 28.68 & 0.9746 & 0.4858 / 0.4830 & $+0.9782$ & 276.1 \\
 & MiniMax-H3 Ref2VA & 0.7476 & 0.86 & 25.09 & 0.9619 & 0.9118 / 0.9232 & $+0.9989$ & 66.4 \\
\midrule
\multirow{6}{*}{\shortstack[l]{Microfiber Bath Towel\\(\emph{Thin-shell forward})}}
 & Direct I2V 1.3B & 0.0132 & 26.16 & 30.70 & 0.9263 & 0.9893 / 0.9877 & $+0.0661$ & 76.0 \\
 & Direct I2V A14B & 0.2458 & 26.97 & 27.50 & 0.9265 & 0.7267 / 0.7304 & $+0.2334$ & 308.6 \\
 & VACE 1.3B (Canny) & 0.3897 & 5.95 & 9.73 & 0.4872 & 0.6315 / 0.6425 & $+0.9503$ & 79.8 \\
 & VACE 1.3B (RGB) & 0.9671 & 0.37 & 12.50 & 0.5931 & 0.3390 / 0.3395 & $+0.9973$ & 75.0 \\
 & VACE A14B (Canny) & 0.7278 & 1.09 & 20.36 & 0.7059 & 0.6050 / 0.6135 & $+0.9300$ & 275.7 \\
 & VACE A14B (RGB) & 0.7957 & 2.37 & 18.51 & 0.6545 & 0.5245 / 0.5078 & $+0.9073$ & 278.0 \\
\midrule
\multirow{6}{*}{\shortstack[l]{Stacked Knitted Gloves\\(\emph{Layered shell peel})}}
 & Direct I2V 1.3B & 0.3479 & 6.07 & 18.38 & 0.7064 & 0.8757 / 0.8743 & $+1.0000$ & 80.7 \\
 & Direct I2V A14B & 0.3274 & 15.32 & 18.30 & 0.8094 & 0.8648 / 0.8653 & $-0.7941$ & 328.6 \\
 & VACE 1.3B (Canny) & 0.3799 & 9.89 & 9.48 & 0.5710 & 0.8297 / 0.8279 & $+0.7946$ & 78.5 \\
 & VACE 1.3B (RGB) & 0.9338 & 0.44 & 14.14 & 0.7103 & 0.4412 / 0.4525 & $+0.9904$ & 81.2 \\
 & VACE A14B (Canny) & 0.5749 & 5.24 & 21.34 & 0.7591 & 0.8814 / 0.8877 & $+0.8581$ & 310.7 \\
 & VACE A14B (RGB) & 0.5921 & 2.28 & 20.97 & 0.7540 & 0.8851 / 0.8913 & $+0.9791$ & 306.4 \\
\midrule
\multirow{6}{*}{\shortstack[l]{Kitchen Sponges \& Bottle\\(\emph{3D hybrid contact})}}
 & Direct I2V 1.3B & 0.2615 & 17.10 & 24.21 & 0.8538 & 0.9816 / 0.9736 & $+0.5269$ & 70.5 \\
 & Direct I2V A14B & 0.5158 & 8.05 & 24.66 & 0.9499 & 0.9199 / 0.9286 & $-0.6672$ & 299.2 \\
 & VACE 1.3B (Canny) & 0.8181 & 1.60 & 14.93 & 0.7251 & 0.9367 / 0.9331 & $+0.9998$ & 70.8 \\
 & VACE 1.3B (RGB) & 0.8720 & 1.44 & 14.83 & 0.7513 & 0.8235 / 0.7929 & $+0.7557$ & 69.6 \\
 & VACE A14B (Canny) & 0.7361 & 3.19 & 19.10 & 0.8058 & 0.9421 / 0.9340 & $+0.9999$ & 283.6 \\
 & VACE A14B (RGB) & 0.7903 & 1.23 & 18.51 & 0.7969 & 0.9461 / 0.9331 & $+0.9947$ & 279.7 \\
\midrule
\multirow{6}{*}{\shortstack[l]{Seat Cushion \& Chair\\(\emph{3D elastic rebound})}}
 & Direct I2V 1.3B & 0.0241 & 34.42 & 29.07 & 0.9518 & 0.9961 / 0.9913 & $-0.9408$ & 83.4 \\
 & Direct I2V A14B & 0.0900 & 16.26 & 20.60 & 0.9150 & 0.9304 / 0.8505 & $+0.4334$ & 320.3 \\
 & VACE 1.3B (Canny) & 0.3762 & 1.91 & 9.06 & 0.4500 & 0.9573 / 0.9377 & $+0.0852$ & 80.5 \\
 & VACE 1.3B (RGB) & 0.7669 & 0.75 & 14.43 & 0.6211 & 0.9557 / 0.9193 & $+0.9415$ & 83.1 \\
 & VACE A14B (Canny) & 0.2247 & 5.67 & 12.85 & 0.5474 & 0.9174 / 0.8705 & --- & 317.1 \\
 & VACE A14B (RGB) & 0.2228 & 4.89 & 13.55 & 0.5724 & 0.9252 / 0.8842 & $-0.2653$ & 304.6 \\
\midrule
\multirow{6}{*}{\shortstack[l]{Microfiber Bath Towel\\(\emph{Counterfactual opposite})}}
 & Direct I2V 1.3B & 0.0043 & 34.30 & 30.15 & 0.9164 & 0.9900 / 0.9878 & $-0.7441$ & 69.7 \\
 & Direct I2V A14B & 0.4444 & 18.51 & 25.49 & 0.9012 & 0.7890 / 0.7898 & $-0.1702$ & 347.9 \\
 & VACE 1.3B (Canny) & 0.3452 & 16.56 & 9.49 & 0.4369 & 0.7028 / 0.6956 & $+0.9930$ & 69.0 \\
 & VACE 1.3B (RGB) & 0.9697 & 0.19 & 12.00 & 0.5828 & 0.3302 / 0.3268 & $+0.9944$ & 69.4 \\
 & VACE A14B (Canny) & 0.5713 & 13.30 & 17.15 & 0.6813 & 0.5130 / 0.5214 & $+0.9579$ & 356.4 \\
 & VACE A14B (RGB) & 0.5680 & 10.34 & 16.75 & 0.6355 & 0.5265 / 0.5305 & $+0.9083$ & 337.7 \\
\midrule
\multirow{2}{*}{Ceramic Mug (\emph{Counterfactual})}
 & Direct I2V 1.3B & 0.0082 & 20.84 & 32.12 & 0.9690 & 0.9890 / 0.9660 & $+0.1234$ & 79.4 \\
 & Direct I2V A14B & 0.2233 & 15.80 & 22.03 & 0.9590 & 0.8883 / 0.8301 & $+0.3867$ & 316.5 \\
\bottomrule
\end{tabular*}
\end{table}

\FloatBarrier
\bibliographystyle{plain}
\bibliography{references}

\end{document}